\documentclass[10pt,twocolumn,letterpaper]{article}

\usepackage{iccv}              
\usepackage{adjustbox}
\usepackage[accsupp]{axessibility}
  
\newcommand{\ours}{CryoFastAR\xspace}

\newcommand{\ab}{\textit{ab initio}\xspace}

\definecolor{iccvblue}{rgb}{0.21,0.49,0.74}
\usepackage[pagebackref,breaklinks,colorlinks,allcolors=iccvblue]{hyperref}
\DeclareMathOperator*{\argmin}{argmin} 

\def\paperID{****} 
\def\confName{ICCV}
\def\confYear{2025}

\title{CryoFastAR: Fast Cryo-EM \textit{Ab initio} Reconstruction Made Easy}

\author{Jiakai Zhang$^{1,2}$
 Shouchen Zhou$^{1,2}$
 Haizhao Dai$^{1,2}$ Xinhang Liu$^{3}$\\
Peihao Wang$^{4}$ Zhiwen Fan$^{4}$ Yuan Pei$^{1}$ Jingyi Yu$^{1}$ \\
\text{$^{1}$ShanghaiTech University} \text{$^{2}$Cellverse, Co., Ltd} \text{$^{3}$HKUST} \text{$^{4}$UT Austin}\\
{\tt\small\{zhangjk,zhoushch,daihzh2023,peiyuan,yujingyi\}@shanghaitech.edu.cn}\\
{\tt\small xliufe@connect.ust.hk}
{\tt\small \{peihaowang, zhiwenfan\}@utexas.edu}}

\begin{document}

\twocolumn[{%
    \renewcommand\twocolumn[1][]{#1}%
    \maketitle
    \centering
    \vspace{-0.8cm}
    \includegraphics[width=0.9\linewidth]{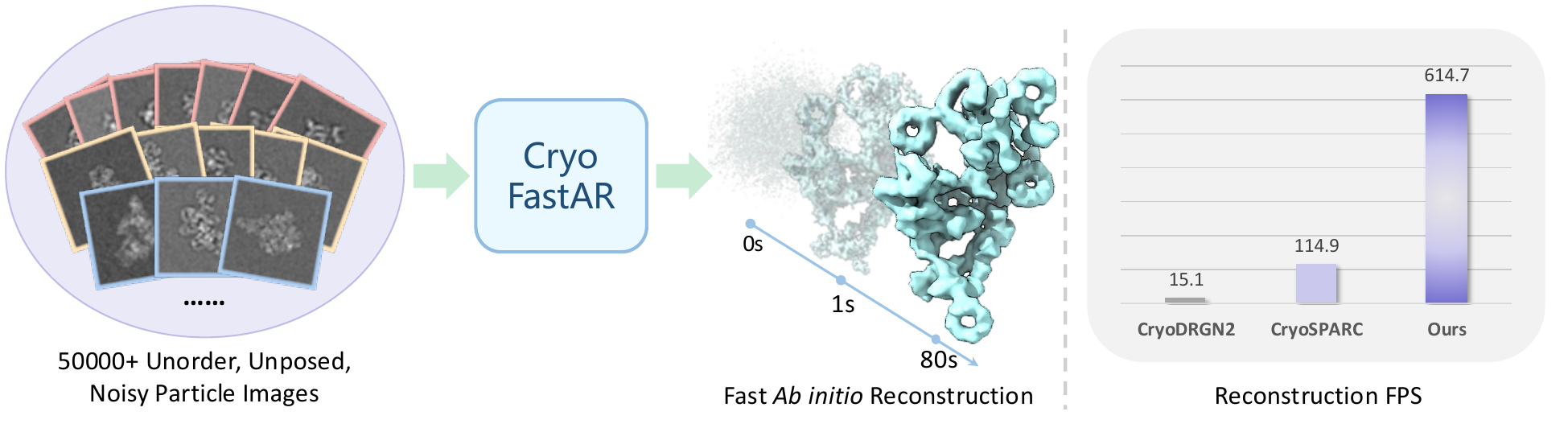}
    \captionof{figure}{\textbf{\ours} enables fast feed-forward \ab reconstruction from hundreds of thousands of unordered, unposed, and highly noisy cryo-EM particle images. Compared to existing baselines, it achieves significantly higher reconstruction speed. We define reconstruction FPS as the average number of particle images processed per second during \ab reconstruction.}
    \vspace{0.4cm}
}]

\begin{abstract}
Pose estimation from unordered images is fundamental for 3D reconstruction, robotics, and scientific imaging. Recent geometric foundation models, such as DUSt3R, enable end-to-end dense 3D reconstruction but remain underexplored in scientific imaging fields like cryo-electron microscopy (cryo-EM) for near-atomic protein reconstruction. In cryo-EM, pose estimation and 3D reconstruction from unordered particle images still depend on time-consuming iterative optimization, primarily due to challenges such as low signal-to-noise ratios (SNR) and distortions from the contrast transfer function (CTF). We introduce CryoFastAR, the first geometric foundation model that can directly predict poses from Cryo-EM noisy images for fast, feed-forward ab initio reconstruction. By integrating multi-view features and training on large-scale simulated cryo-EM data with realistic noise and CTF modulations, \ours enhances pose estimation accuracy and generalization. To enhance training stability, we propose a progressive training strategy that first allows the model to extract essential features under simpler conditions before gradually increasing difficulty to improve robustness. Experiments show that \ours achieves comparable quality while significantly accelerating inference over traditional iterative approaches on both synthetic and real datasets.

\end{abstract}    
\section{Introduction}
\label{sec:intro}

Pose estimation from unordered or sequential images is a cornerstone of numerous disciplines including 3D reconstruction~\cite{3dgs}, robotics, and scientific imaging~\cite{punjani2017cryosparc, sart}. For unordered images without spatiotemporal constraints, pose estimation relies solely on visual consistency across views, making it particularly challenging. Traditional methods have long approached this problem through a per-scene iterative optimization pipeline including feature matching~\cite{bay2006surf}, Structure-from-Motion (SfM)~\cite{hartley2003multiple} and multiple refinement steps to optimize camera poses. While effective, these pipelines suffer from high computational costs and are prone to suboptimal solutions, limiting their scalability. 

Recent geometric foundation models, such as DUSt3R~\cite{wang2024dust3r} are transforming this paradigm toward a generalized per-image inference strategy, enabling end-to-end reconstruction from unposed images, significantly improving performance for large-scale 3D reconstruction tasks. Nevertheless, these advancements remain largely unexplored in scientific imaging, particularly in cryo-electron microscopy (cryo-EM), where accurate pose estimation is fundamental to high-resolution protein reconstruction. In cryo-EM, jointly estimating poses and reconstructing 3D structures from hundreds of thousands of unordered particle images, a process known as \ab reconstruction, still relies on per-scene iterative optimization. This is primarily due to several unique challenges in cryo-EM, including extremely low signal-to-noise ratios (SNR)\cite{cryo-em-formation} and contrast transfer function (CTF) distortions\cite{ctffind4}.

These challenges have typically been addressed using iterative optimization methods per target molecule. Traditional methods such as RELION~\cite{scheres2012relion} and CryoSPARC~\cite{punjani2017cryosparc} adopt an Expectation-Maximization algorithm for \textit{maximum a posteriori} estimation to search 5D pose parameters for every image. More recent approaches, such as CryoAI~\cite{cryoai} and CryoSPIN~\cite{cryospin}, employ image encoders to directly predict image poses, alleviating the need for exhaustive pose searches. However, their performance is often suboptimal due to the non-convex nature of the objective function. To improve the stability as well as the performance of the reconstruction, CryoDRGN2~\cite{cryodrgn2} proposes a hybrid pipeline that iteratively searches poses and conducts a neural reconstruction. Nevertheless, all these methods still require extensive per-image iterative refinement from random initialization and often converge to local minima without careful hyperparameter tuning.

\begin{figure*}[t]
    \centering
    \includegraphics[width=\linewidth]{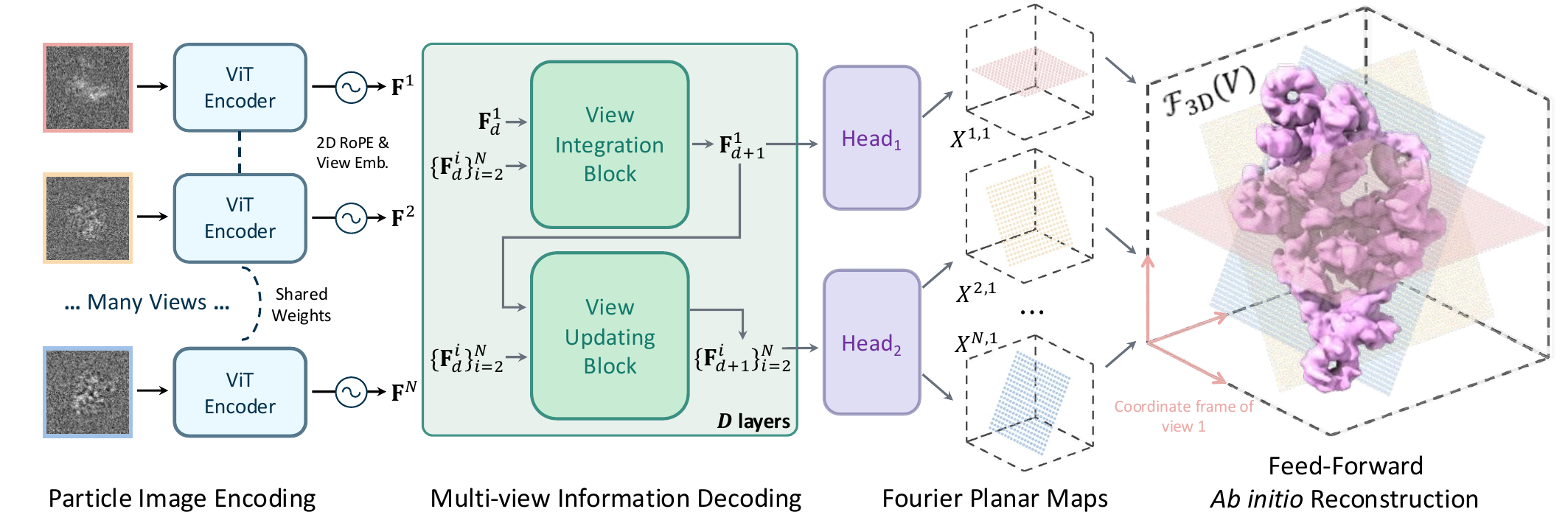}
    \caption{\textbf{Pipeline of \ours.} Our method takes multiple noisy cryo-EM particle images as input and extracts patch-level features using a shared Vision Transformer (ViT) encoder, which incorporates 2D Rotary Position Embeddings (RoPE) and view embeddings. These extracted features are subsequently integrated through stacked View Integration and Refinement blocks. The model outputs Fourier planar maps via two prediction heads, encoding the relative poses of each view with respect to a reference view. Finally, these planar maps are converted to explicit pose parameters, enabling efficient 3D reconstruction via a direct back projection in Fourier space.}
    \label{fig:overview}
    \vspace{-0.2in}
\end{figure*}

To address this, we present \ours, the first geometric foundation model that can directly predict poses from \textbf{Cryo}-EM unordered noisy images for \textbf{Fast} \textit{\textbf{A}b initio} \textbf{R}econstruction. Our method employs a vision transformer-based architecture to directly estimate relative poses in a global coordinate system by efficiently extracting and integrating multi-view image features. Instead of explicitly regressing conventional 5D pose parameters, \ours predicts a dense Fourier planar map to encode each image’s pose relative to a reference view, enabling more flexible optimization. This representation can be readily converted into conventional 5D pose parameters for seamless integration with existing reconstruction algorithms. Additionally, by training on large-scale, realistically simulated cryo-EM data, \ours achieves robust pose estimation without requiring precomputed contrast transfer function (CTF) parameters, simplifying the reconstruction pipeline and enhancing generalization capability.

Our model is trained in a fully supervised manner to predict relative poses from multi-view cryo-EM images. We curate a large-scale cryo-EM simulation dataset consisting of over 110,000 protein structures derived from publicly available PDB models~\cite{pdb}, with each structure associated with 100 simulated images and corresponding ground-truth poses. The diversity of these structures provides extensive geometric variation essential for robust training. To enhance generalization to real data, we apply realistic data augmentations, including random noise addition, CTF simulation, and other perturbations representative of experimental conditions. Moreover, we adopt a progressive training strategy, initially training on clean images with fewer views, then gradually increasing noise levels, CTF distortions, and the number of input views. This staged approach enables the model to first learn essential structural features and subsequently adapt to more challenging data distributions. Finally, we fine-tune the model using a small set of real cryo-EM images of complex proteins, incorporating the maximum number of views to further improve performance on real datasets.

We extensively evaluate \ours on both unseen synthetic and experimental cryo-EM datasets. On synthetic datasets, \ours achieves state-of-the-art results in terms of 2D in-plane shift estimation and overall reconstruction quality. With a rapid pose refinement step in CryoSPARC~\cite{punjani2017cryosparc}, our method achieves superior performance at substantially lower computational cost, accelerating reconstruction by over an order of magnitude (10×) compared to existing methods. On real datasets, \ours demonstrates reconstruction results comparable to current state-of-the-art methods, yet significantly reduces computational time. 

\section{Related Work}
\label{sec:related_work}

\noindent\textbf{\textit{Ab initio} cryo-EM reconstruction} is known to confront significant challenges such as unknown particle poses and extremely low signal-to-noise ratios (SNR).
Early approaches~\cite{elmlund2012simple, scheres2007disentangling} for cryo-EM reconstruction relied on expectation-maximization, which was popularized by RELION~\cite{scheres2007disentangling}. On the other hand, common line-based approaches~\cite{singer2010detecting,pragier2019common,greenberg2017common} select optimal poses by common-line detection. 
Subsequently, stochastic gradient-based optimization techniques~\cite{brubaker2015building} were introduced and adopted in CryoSPARC software~\cite{punjani2017cryosparc}. 
These traditional algorithms begin with a randomly initialized 3D density volume and then determine the best matching pose for each particle image iteratively. All methods reconstruct 3D densities using interpolation and averaging of particle images via the Fourier Slice Theorem (FST)~\cite{hsieh2003computed}.

\noindent\textbf{Neural Representations for Cryo-EM Reconstruction.}
Neural representations have recently emerged as promising tools for addressing continuous structural heterogeneity in cryo-EM reconstruction~\cite{cryodrgn, cryoai, cryofire, ResMFN, cryostar, OPUS-DSD2, cryobench, cryoformer, cryopose}. CryoDRGN~\cite{cryodrgn} introduced a variational autoencoder (VAE)  architecture that encodes conformational states from images and decodes them through a coordinate-based multi-layer perceptrons (MLP) representing the 3D Fourier volume. However, this approach still relies on traditional pipelines for pose estimation and assumes a static structure. More recent neural approaches~\cite{cryodrgn2, cryoai, cryospin} attempt to jointly optimize pose estimation and neural volume reconstruction. CryoDRGN2~\cite{cryodrgn2} and DRGN-AI~\cite{drgnai} integrates iterative pose optimization with neural reconstruction, while CryoAI~\cite{cryoai} employs amortized inference for direct pose estimation, and CryoSPIN~\cite{cryospin} further improves this with a semi-amortized strategy. Nevertheless, all these methods remain limited by computationally intensive per-scene optimization, often resulting in slow convergence and susceptibility to local minima. In contrast, our approach directly predicts Fourier planar maps in a fully feed-forward manner from multi-view particle images, enabling more efficient and accurate reconstructions.

\noindent\textbf{Macroscopic 3D reconstruction: From SfM to DUSt3R.}
The traditional Structure-from-Motion (SfM) pipeline~\cite{hartley2003multiple, schonberger2016structure} decomposes the reconstruction task into sequential subproblems: feature matching~\cite{bay2006surf,lowe2004distinctive}, essential matrix estimation, sparse triangulation, camera pose estimation, and dense reconstruction. While effective, this pipeline accumulates errors at each stage, and its individual subproblems cannot be perfectly addressed by iterative optimization alone. Recent enhancements integrate learning-based techniques for feature description~\cite{detone2018superpoint, dusmanu2019d2} and image matching~\cite{sarlin2019superglue}, yet the fundamental multi-stage structure remains. To overcome these limitations, DUSt3R~\cite{wang2024dust3r} proposes an end-to-end geometric foundation model, directly estimating dense 3D pointmaps from image pairs. MonST3R~\cite{zhang2024monst3r} further extends DUSt3R to monocular settings, predicting pointmaps from single images, while CUT3R~\cite{zhang2024monst3r} generalizes this approach to multi-view scenarios. Our work introduces a similar paradigm shift into cryo-EM, replacing traditional iterative per-image optimization with an efficient feed-forward reconstruction pipeline in microscopic 3D reconstruction.
\section{Preliminary}
\label{sec:prelim}

\subsection{Image Formation Model}

Cryo-EM image formation can be modeled by assuming each observed particle image is generated from an unknown underlying 3D electron density map $V: \mathbb{R}^3 \mapsto \mathbb{R}$.  Each particle undergoes an unknown rotation $R \in \text{SO}(3)$ and an in-plane translation $\mathbf{t} = (t_x, t_y)^\top\in\mathbb{R}^2$ in the camera coordinates, shifting the projected image by $t_x$ pixels along the $x$ axis and $t_y$ pixels along the $y$ axis, respectively. The projection operator $\mathcal{P}_{R,\mathbf{t}}$ integrates the density along the optical axis (the $z$-axis), formally defined as:  
\begin{equation}
(\mathcal{P}_{R,\mathbf{t}} \circ V)(x, y) = \int_{\mathbb{R}} V\left(R\mathbf{p} + h(\mathbf{t})\right)\, \mathrm{d}z,
\end{equation}
where $\mathbf{p} = (x,y,z)^\top$ and $h: \mathbb{R}^2\mapsto\mathbb{R}^3$ is the homogeneous coordinate mapping given by $h(\mathbf{t})=\left(t_x,t_y,0\right)^{\top}$.

Subsequently, each projected 2D image $I^i$ is convolved with the microscope’s point spread function (PSF), which captures lens-induced signal distortions. Finally, additive Gaussian noise~\cite{cryo-em-formation}, a widely adopted noise model in cryo-EM~\cite{cryo-em-formation}, corrupts the image, yielding the observed particle image as:
\begin{equation}
I^i(x, y) = \left[\text{PSF}_i \star (\mathcal{P}_{R_i,\mathbf{t}_i} \circ V)\right](x, y) + \epsilon_i(x, y),
\label{eq:image_formation_real}
\end{equation}
where $\star$ is the convolution operator.

\subsection{Fourier Slicing Theorem}
Most existing cryo-EM reconstruction methods~\cite{punjani2017cryosparc, cryodrgn} rely on the Fourier Slice Theorem (FST)\cite{hsieh2003computed}, as it enables efficient reconstruction of the 3D electron density in Fourier space directly from multiple 2D projections.
Specifically, the 3D Fourier transform of the underlying volume, denoted as $\hat{V} = \mathcal{F}_{\text{3D}}(V)$ can be reconstructed by integrating the 2D Fourier transforms of multiple projection images. Formally, the slicing operation of the volume given a pose can be defined as: 
\begin{equation}
\begin{aligned}
    (\hat{\mathcal{P}}_{R,\mathbf{t}}\circ\hat{V})(\omega_x,\omega_y) &= \hat{V}\left(R(\omega_x,\omega_y,0)^{\top}\right) \\
    &\cdot\exp\left[2\pi j (\omega_x,\omega_y,0)R^{\top}h(\mathbf{t})\right].
\end{aligned}
\end{equation}
In the Fourier domain, the image formation model can be succinctly expressed as:
\begin{equation}
    \hat{I}^i(\omega_x, \omega_y)=\text{CTF}_i\odot(\hat{\mathcal{P}}_{R_i,\mathbf{t}_i}\circ\hat{V})(\omega_x,\omega_y) + \hat{\epsilon}_i(\omega_x, \omega_y),
\end{equation}
where $\text{CTF}_i$ is the microscope's contrast transfer function, which is the 2D Fourier transform of its corresponding $\text{PSF}_i$, and $\hat\epsilon_i$ denotes additive Gaussian noise in the frequency domain. The operator $\odot$ represents element-wise multiplication.

\subsection{Homogeneous Reconstruction}

Once image orientations $R_i$ and in-plane translations $\mathbf{t}_i$ as well as the CTFs have been estimated, the reconstruction is performed in the frequency domain. In this process, the Fourier-transformed images are mapped to a 3D Fourier volume $\hat{V}$ defined over the frequency coordinate $\boldsymbol{\omega} = (\omega_x, \omega_y, \omega_z)^\top$. Specifically, the Fourier volume is computed as
\begin{equation}
\label{eq:reconstruction}
\hat{V}(\boldsymbol{\omega}) = \frac{\sum_{\mathbf{k}_i \in \mathcal{K}(\boldsymbol{\omega})} \operatorname{CTF}_i^*\!\left(\mathbf{k}_i\right) \, \hat{I}^i\!\left(\mathbf{k}_i\right) \, e^{-2 \pi j \mathbf{k}_i^{\top} \mathbf{t}_i}}{\sum_{\mathbf{k}_i \in \mathcal{K}(\boldsymbol{\omega})}\left|\operatorname{CTF}_i\!\left(\mathbf{k}_i\right)\right|^2},
\end{equation}
where the set
$\mathcal{K}(\boldsymbol{\omega}) = \left\{ \mathbf{k}_i = h^{-1}\left(R_i^{\top}\boldsymbol{\omega}\right) \ \big| \ \mathbf{r}_{i}^{\top}\boldsymbol{\omega}=0 \right\}$
collects the relevant Fourier coordinates. Here, $\mathbf{r}_{i}$ denotes the third column vector of $R_i$, and the inverse homogeneous operator \(h^{-1}: \mathbb{R}^3 \mapsto \mathbb{R}^2\) is defined as
$h^{-1}(\boldsymbol{\omega}) = \left(\omega_x, \omega_y\right)^{\top}.$
Finally, the inverse 3D Fourier transform is applied to obtain the real-space volume: $V = \mathcal{F}^{-1}_{\text{3D}}(\hat{V}).$

\section{Methods}
\label{sec:methods}

As shown in Figure~\ref{fig:overview}, \ours employs a standard vision transformer (ViT) architecture that efficiently encodes particle image features and decodes them into multi-view Fourier planar maps. Specifically, a shared ViT encoder extracts patch-level features, which are then refined through stacked view integration blocks to produce informative representations. These representations are subsequently decoded to Fourier planar maps that encode relative poses (Section~\ref{sec:model_architecture}). To effectively train our model on large-scale datasets with high noise levels, we adopt a progressive training strategy, gradually increasing data complexity and noise levels (Section~\ref{sec:training}). During inference, the predicted Fourier planar maps are regressed to explicit 5D poses, facilitating fast \ab protein reconstruction (Section~\ref{sec:inference}).

\paragraph{Fourier planar map.} A fundamental challenge in protein structure recovery is accurately predicting the orientation of a slicing plane in the canonical Fourier space. To address this, we introduce a novel representation termed the \textit{Fourier planar map}, which encodes per-pixel 3D displacements in Fourier space. These displacements indicate where each 2D Fourier-transformed image lies, in accordance with the Fourier slice theorem. Given a 5D pose $(R, \mathbf{t})$, the Fourier planar map is defined as:
\begin{equation}
    X = RX^0 + h(\mathbf{t}), X\in \mathbb{R}^{H \times W \times 3},
\end{equation} where $X^{0}$ represents a uniformly sampled 2D grid on the plane $z = 0$, spanning the range $[-1, 1]^2$. Our goal is to develop a neural network that directly takes a set of cryo-EM images as input and predicts the corresponding dense Fourier planar map. Note that the map does not represent actual Fourier coefficients.

\subsection{Model Architecture}
\label{sec:model_architecture}
\paragraph{Particle image encoding.} Given $N$ multi-view images $\{I^i\}_{i=1}^N$, we first encode each image into patch-wise features $\mathbf{F}^i$  using a ViT-based encoder: 
\begin{equation}
    \mathbf{F}^i = \text{Encoder}(I^i).
\end{equation}
Before these features are passed into the decoder, we apply 2D rotary positional embeddings (2D RoPE)~\cite{croco,croco_v2} to encode spatial positions of patches and introduce learnable, high-dimensional view embeddings to distinguish between different views.

\paragraph{Multi-view information decoding.} Directly applying self-attention across all views introduces quadratic memory and computational complexity, becoming impractical for scenarios involving dozens or hundreds of particle images. To address this, we propose an efficient cross-attention-based approach whose complexity scales linearly with the number of views. Specifically, each decoder layer consists of two key modules: 1) a \textit{view integration block}, which aggregates target views' features into a primary reference view (we choose the first view during the training) via cross-attention; and 2) a \textit{view updating block}, which further refines auxiliary views conditioned on the updated primary features. 

\begin{equation}
\begin{aligned}
    \mathbf{F}^{1}_{d} &= \text{IntBlock}_{d}(\mathbf{F}^1_{d-1}, \{\mathbf{F}^{i}_{d-1}\}_{i=2}^{N}), \\
        \{\mathbf{F}^{i}_{d}\}_{i=2}^{N} &= \text{UpdateBlock}_{d}
    (\mathbf{F}^1_{d}, \{\mathbf{F}^{i}_{d-1}\}_{i=2}^{N}),
\end{aligned}
\end{equation}
where $d = 1, 2, \cdots, D$ for a decoder with $D$ paired integration and update blocks, $\mathbf{F}^i_d$ represents the output feature of $i$-th view's at layer $d$ and the initial feature $\mathbf{F}_0^i $ is defined as $\mathbf{F}^i$. 
Stacking multiple decoder layers progressively integrates multi-view information, resulting in informative representations.

\paragraph{Downstream heads.} 

Finally, the updated feature representation of each view is decoded into a pixel-wise Fourier planar map accompanied by an auxiliary confidence map, expressed within the reference (first) view's 3D Fourier coordinate frame. Specifically, two separate downstream heads perform this decoding: one head predicts the reference view's planar map and the other predicts planar maps for all target views. Formally expressed as:

\begin{equation}
\begin{aligned}
    X^{1,1}, C^{1,1} &= \text{Head}_1(\mathbf{F}^1_{D}), \\  X^{i,1}, C^{i,1} &= \text{Head}_2(\mathbf{F}^i_{D}),
\end{aligned}
\end{equation}
where $X^{i,j}$ is the predicted 3D Fourier planar map of the image $I^j$ in the image $I^i$'s coordinates in the Fourier domain and $C^{i,j}$ is the corresponding confidence map.

\subsection{Training}
\label{sec:training}
\paragraph{Training Objectives.} Our model is trained in a fully supervised manner. Given a set of images $\{I^i\}^{N}_{i=1}$ along with their ground truth orientations $\{R_i\}^{N}_{i=1}$ and in-plane 2D shifts $\{\mathbf{t}_i\}^{N}_{i=1}$, the ground-truth relative Fourier planar map from view $i$ to view $1$, denoted as $\bar{X}^{i, 1} \in \mathbb{R}^{H \times W \times 3}$, can be computed as: 
\begin{equation} 
    \bar{X}^{i, 1} = R_i R_1^{-1} X^0 + h(\mathbf{t}_i).
\end{equation} 
The training objective is a confidence-weighted 3D regression loss:
\begin{equation}
    \mathcal{L}_{3D} = \sum_{i=1}^{N} C^{i,1}\left\| \bar{X}^{i, 1} - {X}^{i, 1} \right\|^2 - \alpha\log C^{i,1},
\end{equation}
where $C^{i,1}$ is activated by $\exp(\cdot) + 1$ to encourage the model to minimize 3D errors in low-confidence regions rather than outputting zero-confidence maps. At inference, the confidence map can effectively highlight regions where predicted Fourier planar maps deviate significantly from their true counterparts.

\paragraph{Progressive Training Scheme.} Directly training the model end-to-end on high-noise cryo-EM images poses significant convergence challenges due to the significant differences between cryo-EM data and typical computer vision datasets. To overcome this, we adopt a progressive training approach consisting of three stages. First, we pre-train our model on clean projection images using only two views per particle, facilitating rapid convergence on simplified conditions. Next, we progressively expand training to our full-scale dataset, gradually increasing input views from 2 to 32 and simultaneously reducing the SNR from 10.0 to 0.1, with added CTF distortions. Finally, to mitigate underfitting on experimental data, we fine-tune our model on a small set of real cryo-EM images. This progressive training strategy ensures stable convergence, enabling the model to robustly extract and integrate multi-view features for accurate \ab reconstruction.

\subsection{Inference}
\label{sec:inference}

Although initially trained with 32 views, we observed that performance improves with more input views, indicating the model's ability to integrate multi-view information. During inference, we use 128 views, fixing one as the reference and predicting Fourier planar maps for the remaining views in batches.

\paragraph{Reference View Selection.} Since our model predicts relative poses with respect to a reference view, it is crucial to ensure the reference is of high quality and not a junk particle. We employ a simple yet effective strategy: we sample 64 candidates and select the one with the highest average confidence to other fixed target views as the reference view.

\paragraph{Pose Regression.} Directly using the predicted planar map positions for reconstruction (Equation~\ref{eq:reconstruction}) may introduce subtle positional inaccuracies, degrading reconstruction quality. To resolve this, we first explicitly regress the 5D image poses from the Fourier planar maps, and subsequently perform traditional Fourier-space back-projection for homogeneous reconstruction. Specifically, the predicted 2D translation of view $i$ is computed as the confidence-weighted average over predicted positions:
\begin{equation} 
\mathbf{t}_i = \dfrac{1}{H \times W} \sum_h \sum_w C^{i,1}_{h, w} X^{i,1}_{h,w}. 
\end{equation} 
The relative orientation of view $i$ with respect to the reference view (view $1$) is estimated by solving the weighted least-squares optimization: 
\begin{equation} 
R_i^* = \argmin_R \sum_{h}\sum_{w}C^{1,1}_{h,w} C^{i,1}_{h,w} \| (X^{i,1}_{h,w} - \mathbf{t}_i) - R_i \bar{X}^{1,1}_{h,w}\|^2. 
\end{equation}
This optimization problem can be solved efficiently via singular value decomposition (SVD). Specifically, the weighted Kabsch algorithm~\cite{Kabsch_1976,Kabsch_1978} is a robust solution that provides accurate orientation estimates and is adopted throughout our experiments. Further improvements in robustness can be achieved using additional outlier-rejection methods such as RANSAC~\cite{ransac}.

\section{Experiments}
\label{sec:experiments}

\begin{figure*}[ht]
    \centering
    \includegraphics[width=0.9\linewidth]{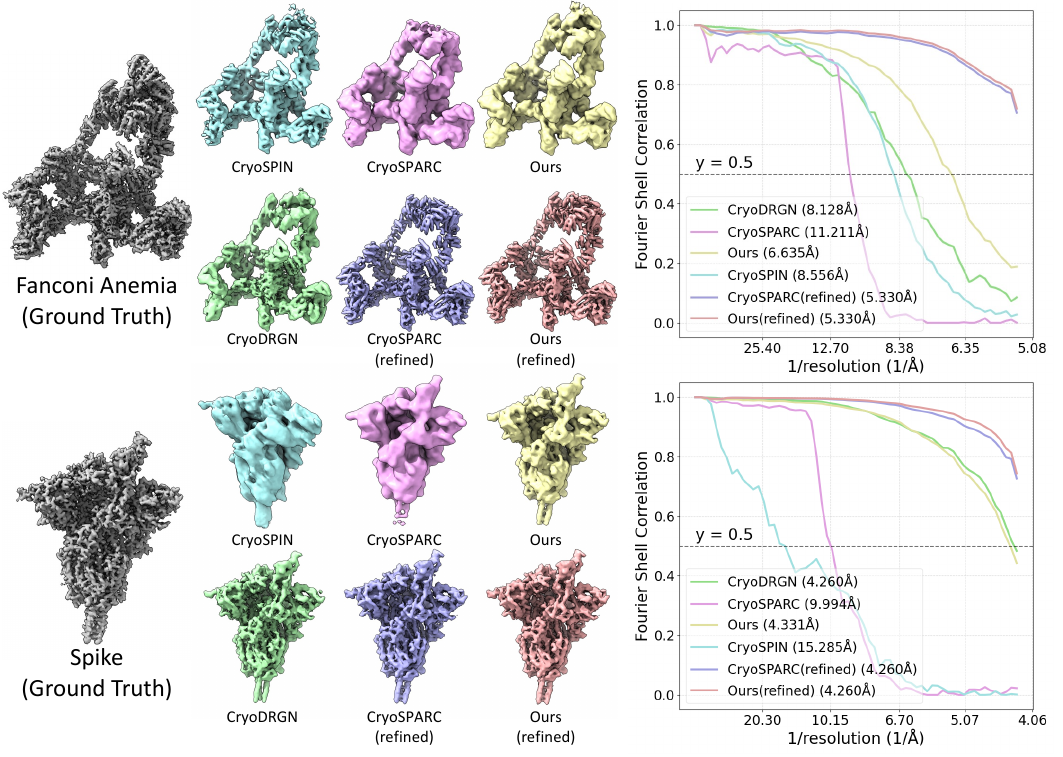}
    \caption{\textbf{Qualitative Results.} We compare our visual quality with all other baselines before and after the refinement for FA and Spike. The results show that our method is comparable to them before refinement and achieves the best performance after the refinement.}
    \label{fig:comparison-synthetic}
    \vspace{-0.1in}
\end{figure*}

\paragraph{Implementation Details.} Our model employs a ViT-Large encoder and a decoder with multiple layers, each composed of a view integration block and a view updating block. Each decoder layer first refines individual view features through a self-attention block, followed by a cross-attention step to integrate multi-view information. Training is conducted progressively in three stages: first, we pre-train the model on clean projection images of a single molecule (PDB ID: 1xvi~\cite{1xvi}) with only two views per particle for 100 epochs to ensure rapid convergence. Next, we expand training to our full-scale simulated dataset for 1000 epochs, gradually increasing the number of input views, and noise levels, and introducing realistic CTF distortions. Finally, we fine-tune the model for another 1000 epochs on real cryo-EM images to bridge the domain gap between simulation and experimental conditions, enhancing generalization to real-world scenarios. In total, we train our model on 32 NVIDIA H20 GPUs for three weeks. All of our experiments are conducted with a single NVIDIA H20 GPU.

\paragraph{Training Dataset.}

To train our model, we construct the first large-scale simulated cryo-EM dataset containing 113,600 atomic structures of protein complexes curated from the Protein Data Bank (PDB). For each atomic structure, we first generate a density volume at a resolution of $128^3$ using EMAN2~\cite{EMAN2}, with voxel sizes automatically determined (ranging from 1Å to 10Å) to fully enclose each structure. From each density volume, we uniformly sample orientations to produce 100 clean 2D projection images at $128\times128$ resolution. To simulate realistic imaging conditions, we further apply data augmentations, including additive random noise, random in-plane 2D shifts, and CTF modulation. The CTF parameters are sampled from 152,385 CTF parameters in CryoCRAB~\cite{cryocrab}. For real-world training, we utilize four experimental datasets from CryoDRGN-EMPIAR~\cite{cryodrgn}, each containing accurate pre-computed particle poses. For each experimental dataset, we randomly select 30,000 images for training and exclude them from evaluation.

\paragraph{Metrics.}

We evaluate our method and baselines using three standard metrics: rotation error, defined as the Frobenius norm of the difference between the ground-truth and estimated rotation matrices; in-plane 2D translation absolute error (in pixels); and reconstruction resolution (measured in \AA). Reconstruction resolution is calculated using the Fourier Shell Correlation (FSC) between the reconstructed and ground-truth volumes, with thresholds of 0.5 for simulated data and 0.143 for experimental data. Additionally, we report the whole evaluation time during the reconstruction. For additional details regarding these metrics, we kindly refer to Appendix C.

\begin{table*}[t]
\caption{\textbf{Quantitative comparison on simulated datasets.} We compare our results with all baselines in terms of rotation error, translation error, and evaluation time. The quantitative results show that our rotation error is comparable to the baselines while achieving the best translation estimation and the fastest inference time. After refinement, our method attains the best overall performance with a slight increase in computation time.}
\label{tab:pose}
\centering
\begin{adjustbox}{max width=\textwidth}
\begin{tabular}{ccccccccccccc}
\hline
Metric & \multicolumn{3}{c}{Rot F-Norm Error, ($\downarrow$)} & \multicolumn{3}{c}{Translation Error (pixel, $\downarrow$)} & \multicolumn{3}{c}{Resolution (\AA, $\downarrow$)} & \multicolumn{3}{c}{Time (mm:ss, $\downarrow$)}\\ \hline
Dataset  & Spliceosome(Sim) & Spike & FA & Spliceosome(Sim) & Spike & FA & Spliceosome(Sim) & Spike & FA & Spliceosome(Sim) & Spike & FA \\ \hline
CryoSPIN & 0.5445 & 1.703 & 0.1855 & - & - & - & 14.67 & 15.29 & 8.56 & 20:51 & 21:30 & 20:49 \\
CryoDRGN2 & 0.0456 & 0.0911 & 0.0679 & 3.5306 & 4.0168 & 5.0338 & 9.50 & \textbf{4.26} & 8.13 & 56:23 & 53:14 & 55:30 \\
CryoSPARC & 0.0501 & 0.0605 & 0.0869 & 1.0035 & 3.8567 & 4.3178 & 10.48 & 9.99 & 11.21 & 11:29 & 04:31 & 05:46 \\
CryoSPARC(refined) & \underline{0.0161} & \underline{0.0283} & \underline{0.0315} & 0.9935 & 0.7202 & 3.6962 & \textbf{8.41} & \textbf{4.26} & \textbf{5.33} & 14:43 & 07:35 & 08:40 \\ 
\hline
Ours & 0.0597 & 0.0416 & 0.0587 & \underline{0.5959} & \underline{0.5469} & \underline{0.7637} & \underline{9.33} & \underline{4.33} & \underline{6.64} & \textbf{01:22} & \textbf{01:21}  & \textbf{01:21} \\
Ours(refined) & \textbf{0.0148} & \textbf{0.0151} & \textbf{0.0169} & \textbf{0.4099} & \textbf{0.4205} & \textbf{0.5074} & \textbf{8.41} & \textbf{4.26} & \textbf{5.33} & \underline{03:37} & \underline{03:42} & \underline{03:53} \\ \hline
\end{tabular}%
\end{adjustbox}

\end{table*}

\begin{table*}[t]
\caption{\textbf{Quantitative comparison on real datasets.} We compare our method with CryoDRGN and CryoSPARC, achieving overall comparable performance in terms of rotation and translation errors. Notably, our method performs particularly well on the RAG and Spliceosome datasets, attaining comparable or superior pose estimation accuracy compared to the baselines.}
\label{tab:comparison-real}
\centering
\begin{adjustbox}{max width=\textwidth}
\begin{tabular}{ccccccccccccc}
\hline
Metric & \multicolumn{3}{c}{Rot F-Norm Error, ($\downarrow$)} & \multicolumn{3}{c}{Translation Error (pixel, $\downarrow$)} & \multicolumn{3}{c}{Time (hh:mm:ss, $\downarrow$)}\\ \hline
Dataset & RAG & 50S & Spliceosome & RAG & 50S & Spliceosome & RAG & 50S & Spliceosome \\ \hline
CryoDRGN2 & 2.1460 & 0.3475 & 2.1698 & 11.6122 & 6.4933 & 15.5078 & 01:32:58 & 01:01:13 & 01:55:55 \\
CryoSPARC & \underline{1.5100} & \underline{0.2110} & 2.3999 & 7.0756 & \underline{1.5993} & 17.4008 & 00:04:44 & 00:10:20 & 00:12:00 \\
CryoSPARC(refined) & \textbf{1.4303} & \textbf{0.1042} & 2.3897 & 6.3806 & \textbf{1.5369} & 18.4135  & 00:10:31 & 00:13:42 & 00:18:10 \\ \hline
Ours & 1.7292 & 1.1521 & \textbf{0.9564} & \underline{5.1064} & 2.1989 & \textbf{4.8698} & \textbf{00:02:39} & \textbf{00:01:58} & \textbf{00:03:31} \\
Ours(refined) & 1.6089 & 0.9355 & \underline{0.9734} & \textbf{4.5414} & 2.2584 & \underline{4.9134} & \underline{00:06:54} & \underline{00:05:47} & \underline{00:08:03} \\ \hline
\end{tabular}%
\end{adjustbox}
\end{table*}


\begin{figure}
    \centering
    \includegraphics[width=\linewidth]{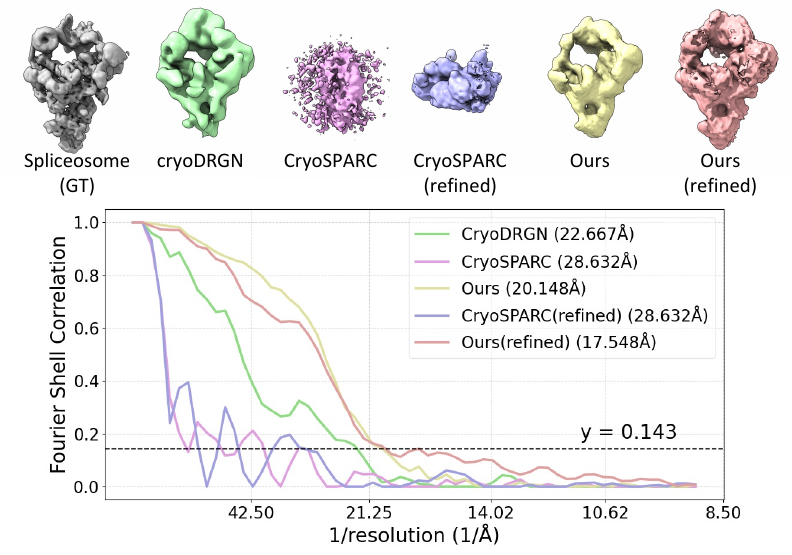}
    \caption{\textbf{Qualitative comparison results on experimental Spliceosome dataset.} Our method achieves the best visual quality and reconstruction resolution compared to other baselines, while CryoSPARC fails to converge to the correct structure due to the heterogeneity of the spliceosome.}
    \label{fig:comparison-real}
    \vspace{-0.2in}
\end{figure}

\subsection{Comparison}

\paragraph{Datasets.} We evaluate our method on both simulated and experimental datasets. For simulated evaluation, we select three representative protein complexes that are widely used benchmarks in cryo-EM: the spliceosome structure (\textbf{Spliceosome(Sim)}, PDB: 5nrl)~\cite{spliceosome1}, a variant of the SARS-CoV-2 spike protein (\textbf{Spike}, PDB: 7sbr)~\cite{spike}, and the human Fanconi anaemia core complex (\textbf{FA}, PDB ID: 7kzp)~\cite{fa}. For experimental evaluation, we test on three datasets from CryoDRGN-EMPIAR~\cite{cryodrgn}: the RAG1-RAG2 complex (\textbf{RAG}, EMPIAR-10049)~\cite{RAG1-RAG2}, 50S Ribosome (\textbf{50S}, EMPIAR-10076) and the pre-catalytic spliceosome (\textbf{Spliceosome}, EMPIAR-10180). More details on these datasets can be found in Appendix B.

\paragraph{Baselines.}
We compare our approach against three \ab reconstruction baselines: (1) \textbf{CryoSPARC}, a widely-used software utilizing iterative expectation-maximization and SGD for \ab reconstruction; (2) \textbf{CryoSPIN}, a neural method directly predicting image poses from single-view inputs; and (3) \textbf{CryoDRGN2}, a hybrid method alternating neural reconstruction and iterative pose search. Notably, unlike these methods, our approach \textbf{does not require pre-computed CTF parameters} for pose estimation. All baseline methods rely on provided CTF parameters during iterative pose refinement or per-scene optimization.
Since the official implementation of CryoSPIN does not support shift estimation, we disable random shifts when training it on simulated data and exclude it from comparisons on experimental data.  After initial \ab reconstruction, we further \textbf{refine} our model’s predicted poses using CryoSPARC's homogeneous refinement to evaluate the quality of our initialization compared to CryoSPARC’s own initialization. Specifically, given our estimated poses, we run CryoSPARC’s reconstruction only followed by the local refinement. For CryoSPARC, we continue to run homogeneous refinement followed by a local refinement.

\paragraph{Simulated results.} 

We compare our method with baseline approaches both qualitatively and quantitatively. As shown in Figure~\ref{fig:comparison-synthetic}, our approach produces reconstructions with superior structural completeness and finer detail preservation. Quantitative results reported in Table~\ref{tab:pose} further confirm that our method outperforms baselines in pose estimation accuracy and reconstruction quality, achieving significantly lower rotation and translation errors. Remarkably, our method accelerates the reconstruction process by over an order of magnitude (10×), while maintaining or surpassing the quality of all baselines. After refinement, our approach consistently achieves the best overall performance, demonstrating \ours's strong initialization quality, computational efficiency, and robustness.

\noindent\textbf{Experimental results.}
Table~\ref{tab:comparison-real} summarizes the quantitative results on three experimental cryo-EM datasets including \textbf{RAG}, \textbf{50S} ribosome, and pre-catalytic \textbf{Spliceosome}~\cite{spliceosome2}. Overall, our approach achieves competitive reconstruction accuracy compared to baseline methods while significantly reducing the computational cost. Specifically, our method achieves approximately a 
$3.33\times$ speed-up compared to CryoSPARC and over a $33.21\times$ speed-up relative to CryoDRGN2. Notably, our method performs particularly well on the RAG and spliceosome datasets, attaining comparable or better pose estimation accuracy relative to the baselines. However, on the 50S ribosome dataset, our method, although predicting poses in generally correct orientations, yields substantially lower accuracy than the baselines. We hypothesize that this discrepancy arises from the intrinsic structural flexibility and complexity associated with membrane proteins like 50S, posing greater challenges for our feed-forward approach trained primarily on simulated globular structures. As demonstrated in Figure~\ref{fig:comparison-real}, we provide visual comparisons of reconstructions on the experimental spliceosome dataset. These visualizations confirm our quantitative comparisons, demonstrating that our refined reconstruction exhibits improved structural fidelity and better captures the detailed molecular features compared to baseline methods. Other baselines may fail due to lacking the robustness of their heterogeneity without direct supervision during optimization. 

\begin{figure}
    \centering
    \includegraphics[width=1.0\linewidth]{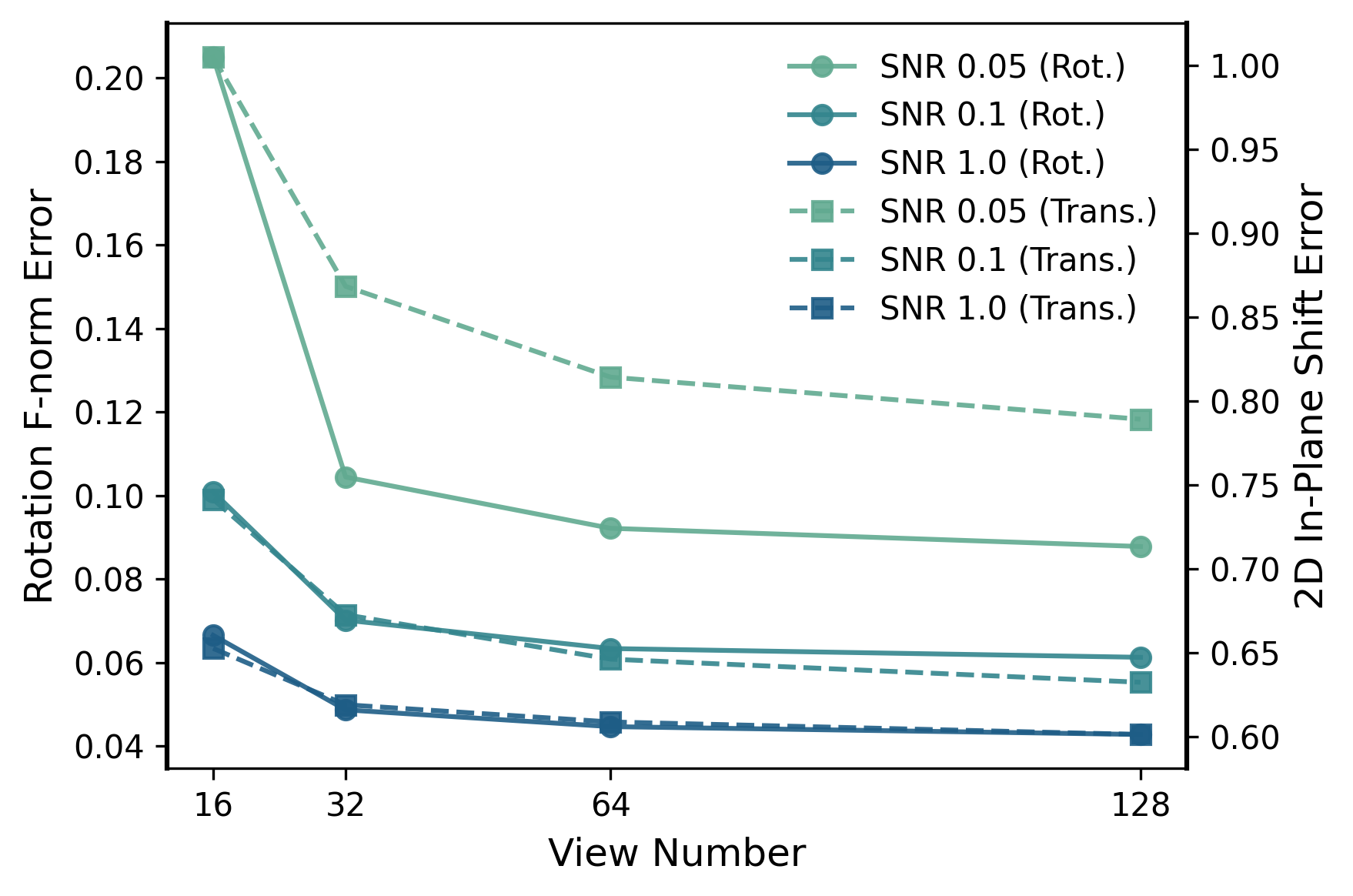}
    \vspace{-0.3in}
    \caption{\textbf{Evaluation on view numbers and SNR.} Our model shows robust performance across different SNRs and achieves better results when the input view number increases.}
    \label{fig:ablation}
    \vspace{-0.2in}
\end{figure}
\subsection{Evaluation}

\paragraph{Effect of view numbers.}
To determine the optimal number of input views, we evaluate performance using 16, 32, 64, and 128 views. As shown in Figure~\ref{fig:ablation}, both the rotation F-norm error and 2D in-plane shift error decrease monotonically as the number of views increases. Specifically, increasing the number of input views from 32 to 128 at SNR 0.1 reduces the rotation F-norm error by 12.6\% and the 2D in-plane translation error by 3.94\%. This trend is consistent across different SNR settings and becomes more pronounced at lower SNRs, which aligns with intuition—noisier images require more input views for robust joint estimation of rotation and 2D shift. These observations highlight our model’s ability to effectively aggregate long sequences of multi-view information.

\paragraph{Effect of SNR.}
As shown in Figure~\ref{fig:ablation}, we evaluate our model’s robustness across different SNR levels, including 0.05, 0.1 (training setting), and 1.0. The results indicate that our model remains effective even when the SNR is reduced to half of the training setting while significantly improving performance at higher SNRs. This suggests that \ours generalizes well across diverse SNR conditions despite being trained on a single SNR level.
\section{Discussion}
\label{sec:conclusion}
\paragraph{Limitations.}
As the first geometric foundation model for direct pose estimation from multi-view noisy cryo-EM images, our method has some limitations. It is mainly trained on simulated data, leading to performance drops on real images due to domain gaps and limited annotations. This can be mitigated with more realistic simulations~\cite{cryogem} or high-quality labeled data. Additionally, it processes only a subset of images per forward pass, limiting reconstruction accuracy. Scalable approaches like Fast3R~\cite{fast3r} and Spann3R~\cite{spann3r} offer promising solutions. 

\paragraph{Conclusion.}
We have introduced \ours, the first geometric foundation model for fast \ab reconstruction in cryo-EM. By leveraging a ViT-based architecture and predicting Fourier planar maps, \ours has effectively integrated multi-view information without requiring iterative pose search. Extensive experiments have demonstrated that our method achieves competitive performance on real and synthetic datasets while significantly reducing computational costs. Our findings have highlighted the potential of feed-forward architectures in cryo-EM reconstruction, paving the way for more efficient and scalable structural analysis. 

\section{Acknowledgment}
This work was supported by ShanghaiTech University’s HPC Platform. We would like to thank the Cellverse team for their valuable discussions.

\clearpage
\begin{center}
\LARGE\textbf{---Supplementary Material---\\
CryoFastAR: Fast Cryo-EM \textit{Ab initio} Reconstruction Made Easy}
\end{center}
\vspace{9pt}
\setcounter{section}{0}
\renewcommand{\thesection}{\Alph{section}}

\begin{figure*}[ht]
    \centering
    \includegraphics[width=\linewidth]{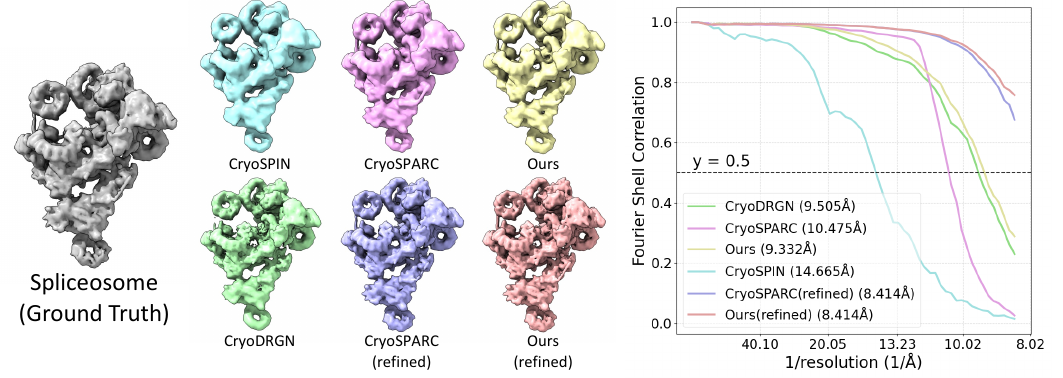}
    \caption{\textbf{Qualitative Result.} We compare our visual quality with all other baselines before and after the refinement for Spliceosome's simulated dataset. The results show that our method is comparable to them before refinement and achieves the best performance after the refinement.}
    \label{fig:additional-comparison}
\end{figure*}

\section{Additional Results}

We compare the visual results of our method with baselines methods on the simulated Spliceosome~\cite{spliceosome1} dataset. The results are shown in Figure~\ref{fig:additional-comparison}, and the quantitative results are presented in Table 1 of the paper.

\section{Details of Dataset}

The generation process for the simulated dataset follows the procedure outlined in Section \ref{sec_supp:dataset}, under the paragraph titled \textbf{Simulated Particle Image Generation}, with some differences in the number of structures used and the number of projections.

\subsection{Simulated dataset.}
We generate 3 simulated datasets for evaluation of the baselines. Each dataset has the same simulation procedure: 1) generating 50000 particles with uniformly sampled rotations, 2) adding CTF corruption 3) translating image in $[-10,10]$ pixels along $x$ and $y$ axes, respectively, 4) add Gaussian noise to adjust the signal-to-noise ratio (SNR) of the image to 0.1. Each structure in the dataset has a different spatial resolution in terms of \AA\ per pixel(Apix) when the PDB~\cite{pdb} structure is converted into a volume density map using EMAN2~\cite{EMAN2}. The PDB ID and Apix for each dataset is as follows:
\begin{itemize}
    \item The spliceosome structure (\textbf{Spliceosome}, PDB ID: 5nrl)~\cite{spliceosome1}. Apix: 4.00.

    \item A variant of the SARS-CoV-2 spike protein (\textbf{Spike}, PDB ID: 7sbr)~\cite{spike}. Apix: 2.03.

    \item the human Fanconi anaemia core complex (\textbf{FA}, PDB ID: 7kzp)~\cite{fa}. Apix: 2.54.
\end{itemize}

\subsection{Experimental dataset.}

For the experimental evaluation, we test on three datasets (EMPIAR-10049, EMPIAR-10076, and EMPIAR-10180) from EMPIAR~\cite{empiar}. After processing these datasets using the scripts provided by CryoDRGN-EMPIAR~\cite{cryodrgn}, we obtain the filtered particle stacks, along with the pre-computed accurate particle poses and 2D in-plane translations. We assume that the reconstructed structures, using the pre-computed poses and translations provided by the datasets, represent the ground-truth volume. The particles are then split into two sets: 30,000 particles for training and the remaining particles for evaluation. The number of images in the evaluation set and the Apix for each dataset are as follows:

\begin{itemize}
    \item The RAG1-RAG2 complex (\textbf{RAG}, EMPIAR-10049)~\cite{RAG1-RAG2}. Apix: 1.845, number particles in the evaluation set: 78544.

    \item The assembling bacterial 50S ribosome (\textbf{50S}, EMPIAR-10076)~\cite{50Sribosome}. Apix: 3.275, number particles in the evaluation set: 57327.

    \item The pre-catalytic spliceosome (\textbf{Spliceosome}, EMPIAR-10180)~\cite{spliceosome2}. Apix: 4.25, number particles in the evaluation set: 109722.
\end{itemize}

\subsection{Contrast Transfer Function}

In cryo-electron microscopy (cryo-EM), the imaging process is influenced by the point spread function (PSF), which characterizes the system's spatial response to a point source. The PSF encapsulates the effects of diffraction, aberrations, and other instrumental imperfections, thereby determining how the interactions between the high-energy electron beam and the specimen are distributed in the final image in real space.

For image processing and analysis, it is often advantageous to work in the frequency domain, typically by applying a Fourier or Hartley transform. In this domain, the Fourier transform of the PSF is referred to as the Contrast Transfer Function (CTF), which describes how different spatial frequency components are modulated by the microscope’s optics.

In our work, instead of applying the PSF directly, we incorporate its effects through the CTF. This approach simplifies the image processing workflow by enabling us to operate entirely in the frequency domain. Following the methodology implemented in CTFFIND4~\cite{ctffind4}, the CTF is defined as:
\begin{equation}
\begin{aligned}
\text{CTF}(w, \lambda, \textbf{g}, \Delta f, C_s, \Delta\varphi) = {} \\
-\sqrt{1-w^2}\sin[\chi(\lambda, |\textbf{g}|, \Delta f, C_s, \Delta \varphi)] \\
-w \cos[\chi(\lambda, |\textbf{g}|, \Delta f, C_s, \Delta\varphi)]
\end{aligned}
\label{equ:ctf}
\end{equation}
where
\begin{equation}
\chi\left(\lambda, |\textbf{g}|, \Delta f, C_s, \Delta \varphi\right) = \pi\lambda|\textbf{g}|^2\left(\Delta f-\frac{1}{2}\lambda^2|\textbf{g}|^2C_s\right)+\Delta\varphi.
\label{equ:ctf_2}
\end{equation}
In these equations, $w$ represents the relative phase contrast factor, while $\chi$ is a frequency-dependent phase shift function. The function $\chi$ incorporates key parameters, including the electron wavelength $\lambda$, the spatial frequency vector $\textbf{g}$, the objective defocus $\Delta f$, the spherical aberration $C_s$, and the phase shift $\Delta\varphi$. The parameters $w$, $\lambda$, $C_s$, and $\Delta\varphi$ are intrinsic to the cryo-EM hardware.

\section{Details of Baselines}
\paragraph{CryoSPIN~\cite{cryospin}.} We use the official implementation of CryoSPIN in \href{https://github.com/shekshaa/semi-amortized-cryoem}{Github}. In our experiments, we run CryoSPIN with its default setting while using our customized datasets as described in the main paper. Also, the official implementation does not include the estimation of the in-plane translation, we omit the random shift when training CryoSPIN. As CryoSPIN often falls to local minima, we run it three times per experiment and report the best result.

\paragraph{CryoDRGN2~\cite{cryodrgn2}.}
We use the official implementation of CryoDRGN v3.4.3 in \href{https://github.com/ml-struct-bio/cryodrgn}{Github}, with all default parameter values, except setting the batch size to 32 while using the `abinit\_homo' command for \ab.

\paragraph{CryoSPARC~\cite{punjani2017cryosparc}.}
We use the CryoSPARC software package v4.6.2. with all default parameter values. We followed the typical workflow: Import particle stacks and then perform \ab reconstruction.

\paragraph{CryoSPARC(refined).}
We use the CryoSPARC software package v4.6.2. with all default parameter values for refinement. We use the result mentioned in paragraph \textbf{CryoSPARC}, continuously performs Homogeneous Refine, and finally a Local Refinement.

\paragraph{Ours(refined).}
We use the CryoSPARC software package v4.6.2 with all default parameter values for refinement. We use Import Particles, and then Reconstruction Only to generate volumes and masks for the future usage, and then run Local Refinement.

\section{Details of Evaluation Metrics}
\label{sec_supp:metrics}
\paragraph{Rotation F-norm error.}
Given a sequence of ground-truth 3D orientations $R_1,\ldots, R_N$ and a sequence of estimated 3D orientations $\hat{R}_1,\ldots,\hat{R}_N$, we randomly select one view, indexed as $i$, to serve as the reference view. To report the F-norm rotation errors for each dataset, we randomly sample 5,000 views as reference views and select the minimum error as the final result. For each random selection $i$, all ground-truth poses are transformed into the coordinate system of this reference view: $R'_{i,j}\gets R_iR_j^{\top},\forall j=\{1,\ldots, N\}$. Similarly, all predicted poses are transformed into the coordinate system of the reference view of the $i$-th predicted pose: $\hat{R}'_{i,j}\gets \hat{R}_i\hat{R}_j^{\top},\forall j=\{1,\ldots, N\}$. The error for each view is then computed by taking the Frobenius norm of the difference between the transformed ground-truth pose and the corresponding transformed predicted pose. Finally, the average of these errors across all views is computed:
\begin{equation}
    \mathcal{L}_{\text{rot}, i} = \dfrac{1}{N}\sum_{j=1}^N\|R'_{i,j} - \hat{R}'_{i,j}\|_F.
\end{equation}

\paragraph{In-plane translation error.}
Given a ground-truth 2D in-plane translation $\mathbf{t} = (t_x, t_y)$ and an estimated translation $\mathbf{\hat{t}} = (\hat{t}_x, \hat{t}_y)$, we compute the mean of the L2-norm:
\begin{equation}
    \mathcal{L}_{2}(\mathbf{t}, \mathbf{\hat{t}}) = \|\mathbf{t} - \mathbf{\hat{t}}\|_2.
\end{equation}
We report the average of the L2 translation errors over each dataset.

\paragraph{Resolution.} 
The reconstruction resolution is calculated using the Fourier Shell Correlation (FSC) between the aligned reconstructed and ground-truth volumes, with thresholds of 0.5 for simulated data and 0.143 for experimental data serving as the metric for pose estimation reconstruction resolution. The formula for FSC is given below:
\begin{equation}
    FSC(r)=\frac{\sum_{r_i\in r}F_1(r_i)\cdot F_2(r_i)^*}{\sqrt{\sum_{r_i\in r} \Vert F_1(r_i)\Vert^2 \cdot \sum_{r_i\in r} \Vert F_2(r_i)\Vert^2}}
    \label{equ:fsc}
\end{equation}
where $F_1, F_2$ are the Fourier transforms of the reconstructed and ground-truth volumes, respectively. $r$ represents all three-dimensional frequency components shown in a one-dimensional form. We use CryoSPARC's Align 3D Maps to automatically align the ground-truth volume with the other reconstructed volume.

\section{Details of Simulated Training Dataset Construction}
\label{sec_supp:dataset}
\paragraph{Data Curation.}
The data curation pipeline for our atomic structure dataset is designed to obtain high-quality and biologically correct 3D structures from the Protein Data Bank(PDB)~\cite{pdb}. The process involves several key steps, including parsing, metadata extraction, filtering, and structural refinement.

\paragraph{Parsing and Metadata Extraction}
The initial step involves parsing the input files in the mmCIF format. The pipeline extracts essential metadata, including the release date, resolution, and experimental method. These metadata are crucial for ensuring the relevance and reliability of the structures included in the dataset.

\paragraph{Filtering Criteria}
The dataset is subjected to stringent filtering criteria to ensure the quality and suitability of the structures for downstream analysis. The filtering process includes:
\begin{itemize}
    \item Release Date: Structures must have been released to the PDB before the cutoff date of 2021-09-30.
    \item Resolution: Only structures with a reported resolution of 9\AA~or less are retained.
    \item Hydrogen Removal: Hydrogen atoms are removed from the structures.
    \item Polymer Chain Integrity: Polymer chains with all unknown residues are removed.
    \item Clashing Chains: Chains with more than 30\% of atoms within 1.7\AA~ of an atom in another chain are identified as clashing. In cases where two chains are clashing, the chain with the greater percentage of clashing atoms is removed. If the same fraction of atoms are clashing, the chain with fewer total atoms is removed. If the chains have the same number of atoms, the chain with the larger chain ID is removed.
    \item Residue and Small Molecule Integrity: For residues or small molecules with CCD codes, atoms outside of the CCD code’s defined set of atom names are removed. Protein chains with consecutive $C\alpha$ atoms larger than 10\AA~ apart are filtered out.
    \item Bioassembly Selection: For bioassemblies with more than 20 chains, a random interface token is selected, ensuring that the center atom is within 15\AA~ of the center atom of a token in another chain.
\end{itemize}
\paragraph{Structural Refinement}
To simplify subsequent analysis, the pipeline performs basic structural cleanup. This includes resolving alternative locations for atoms/residues by selecting the one with the largest occupancy and removing water and crystallization aids. Ligands, peptides, and nucleic acids are also removed to focus on the core protein structures.

\paragraph{Simulated Particle Image Generation}
We begin with the 113,600 curated 3D atomic structures and generate noisy projection images for training through the following steps: For each 3D structure, we convert it into a $128^3$ 3D volume density map using EMAN2~\cite{EMAN2}, the pixel size is set by the maximum length of the structure to make sure the whole structure is inside. Then we uniformly sample 100 projection images in SO(3) space, representing different views of the volume. Each clean projection image is then modified by randomly applying a Contrast Transfer Function (CTF) sampled from real distribution including 152,385 CTF parameters as described in ~\cite{draco}, simulating imaging system imperfections such as aberrations or blurring. A random 2D shift, within the range of $[-10, 10]^2$, is applied to each image to account for potential translational errors during imaging. Finally, Gaussian noise is added to each image to ensure the signal-to-noise ratio (SNR) matches the desired level, simulating the noise characteristics in experimental data. This augmentation pipeline produces diverse, realistic projections for model training.

{
    \small
    \bibliographystyle{ieeenat_fullname}
    \bibliography{main}

\begin{thebibliography}{54}
\providecommand{\natexlab}[1]{#1}
\providecommand{\url}[1]{\texttt{#1}}
\expandafter\ifx\csname urlstyle\endcsname\relax
  \providecommand{\doi}[1]{doi: #1}\else
  \providecommand{\doi}{doi: \begingroup \urlstyle{rm}\Url}\fi

\bibitem[Andersen and Kak(1984)]{sart}
Anders~H Andersen and Avinash~C Kak.
\newblock Simultaneous algebraic reconstruction technique (sart): a superior implementation of the art algorithm.
\newblock \emph{Ultrasonic imaging}, 6\penalty0 (1):\penalty0 81--94, 1984.

\bibitem[Bay et~al.(2006)Bay, Tuytelaars, and Van~Gool]{bay2006surf}
Herbert Bay, Tinne Tuytelaars, and Luc Van~Gool.
\newblock Surf: Speeded up robust features.
\newblock In \emph{Computer Vision--ECCV 2006: 9th European Conference on Computer Vision, Graz, Austria, May 7-13, 2006. Proceedings, Part I 9}, pages 404--417. Springer, 2006.

\bibitem[Berman et~al.(2000)Berman, Westbrook, Feng, Gilliland, Bhat, Weissig, Shindyalov, and Bourne]{pdb}
Helen~M. Berman, John Westbrook, Zukang Feng, Gary Gilliland, T.~N. Bhat, Helge Weissig, Ilya~N. Shindyalov, and Philip~E. Bourne.
\newblock The protein data bank.
\newblock \emph{Nucleic Acids Research}, 28\penalty0 (1):\penalty0 235--242, 2000.

\bibitem[Brubaker et~al.(2015)Brubaker, Punjani, and Fleet]{brubaker2015building}
Marcus~A Brubaker, Ali Punjani, and David~J Fleet.
\newblock Building proteins in a day: Efficient 3d molecular reconstruction.
\newblock In \emph{Proceedings of the IEEE Conference on Computer Vision and Pattern Recognition}, pages 3099--3108, 2015.

\bibitem[Chen et~al.(2025)Chen, Xu, Dai, Shen, Zhang, Liu, Pei, and Yu]{cryocrab}
Qihe Chen, Zhenyang Xu, Haizhao Dai, Yingjun Shen, Jiakai Zhang, Zhijie Liu, Yuan Pei, and Jingyi Yu.
\newblock A large-scale curated and filterable dataset for cryo-em foundation model pre-training.
\newblock \emph{Scientific Data}, 12\penalty0 (1):\penalty0 960, 2025.

\bibitem[Davis et~al.(2016)Davis, Tan, Carragher, Potter, Lyumkis, and Williamson]{50Sribosome}
Joseph~H. Davis, Yong~Zi Tan, Bridget Carragher, Clinton~S. Potter, Dmitry Lyumkis, and James~R. Williamson.
\newblock Modular assembly of the bacterial large ribosomal subunit.
\newblock \emph{Cell}, 167:\penalty0 1610--1622.e15, 2016.

\bibitem[DeTone et~al.(2018)DeTone, Malisiewicz, and Rabinovich]{detone2018superpoint}
Daniel DeTone, Tomasz Malisiewicz, and Andrew Rabinovich.
\newblock Superpoint: Self-supervised interest point detection and description.
\newblock In \emph{Proceedings of the IEEE conference on computer vision and pattern recognition workshops}, pages 224--236, 2018.

\bibitem[Dusmanu et~al.(2019)Dusmanu, Rocco, Pajdla, Pollefeys, Sivic, Torii, and Sattler]{dusmanu2019d2}
Mihai Dusmanu, Ignacio Rocco, Tomas Pajdla, Marc Pollefeys, Josef Sivic, Akihiko Torii, and Torsten Sattler.
\newblock D2-net: A trainable cnn for joint description and detection of local features.
\newblock In \emph{Proceedings of the ieee/cvf conference on computer vision and pattern recognition}, pages 8092--8101, 2019.

\bibitem[Elmlund and Elmlund(2012)]{elmlund2012simple}
Dominika Elmlund and Hans Elmlund.
\newblock Simple: Software for ab initio reconstruction of heterogeneous single-particles.
\newblock \emph{Journal of structural biology}, 180\penalty0 (3):\penalty0 420--427, 2012.

\bibitem[Fischler and Bolles(1981)]{ransac}
Martin~A. Fischler and Robert~C. Bolles.
\newblock Random sample consensus: A paradigm for model fitting with applications to image analysis and automated cartography.
\newblock \emph{Communications of the ACM}, 24\penalty0 (6):\penalty0 381--395, 1981.

\bibitem[Greenberg and Shkolnisky(2017)]{greenberg2017common}
Ido Greenberg and Yoel Shkolnisky.
\newblock Common lines modeling for reference free ab-initio reconstruction in cryo-em.
\newblock \emph{Journal of structural biology}, 200\penalty0 (2):\penalty0 106--117, 2017.

\bibitem[Hartley and Zisserman(2003)]{hartley2003multiple}
Richard Hartley and Andrew Zisserman.
\newblock \emph{Multiple view geometry in computer vision}.
\newblock Cambridge university press, 2003.

\bibitem[Hsieh(2003)]{hsieh2003computed}
J. Hsieh.
\newblock \emph{Computed Tomography: Principles, Design, Artifacts, and Recent Advances}.
\newblock SPIE Press, 2003.

\bibitem[Iudin et~al.(2016)Iudin, Korir, Salavert-Torres, Kleywegt, and Patwardhan]{empiar}
Andrii Iudin, Paul~K. Korir, Jos{\'e} Salavert-Torres, Gerard~J. Kleywegt, and Ardan Patwardhan.
\newblock Empiar: a public archive for raw electron microscopy image data.
\newblock \emph{Nature Methods}, 13\penalty0 (5):\penalty0 387--388, 2016.

\bibitem[Jeon et~al.(2025)Jeon, Raghu, Astore, Woollard, Feathers, Kaz, Hanson, Cossio, and Zhong]{cryobench}
Minkyu Jeon, Rishwanth Raghu, Miro Astore, Geoffrey Woollard, Ryan Feathers, Alkin Kaz, Sonya~M. Hanson, Pilar Cossio, and Ellen~D. Zhong.
\newblock Cryobench: Diverse and challenging datasets for the heterogeneity problem in cryo-em, 2025.

\bibitem[Kabsch(1976)]{Kabsch_1976}
W. Kabsch.
\newblock {A solution for the best rotation to relate two sets of vectors}.
\newblock \emph{Acta Crystallographica Section A}, 32\penalty0 (5):\penalty0 922--923, 1976.

\bibitem[Kabsch(1978)]{Kabsch_1978}
W. Kabsch.
\newblock {A discussion of the solution for the best rotation to relate two sets of vectors}.
\newblock \emph{Acta Crystallographica Section A}, 34\penalty0 (5):\penalty0 827--828, 1978.

\bibitem[Kerbl et~al.(2023)Kerbl, Kopanas, Leimk{\"u}hler, and Drettakis]{3dgs}
Bernhard Kerbl, Georgios Kopanas, Thomas Leimk{\"u}hler, and George Drettakis.
\newblock 3d gaussian splatting for real-time radiance field rendering.
\newblock \emph{ACM Trans. Graph.}, 42\penalty0 (4):\penalty0 139--1, 2023.

\bibitem[Kim et~al.(2002)Kim, Skarina, Beasley, Laskowski, Arrowsmith, Joachimiak, Edwards, and Savchenko]{1xvi}
Y. Kim, Tatiana Skarina, Steven Beasley, Roman~A. Laskowski, Cheryl~H. Arrowsmith, Andrzej~J. Joachimiak, A.~E. Edwards, and Alexei Savchenko.
\newblock Crystal structure of escherichia coli ec1530, a glyoxylate induced protein ygbm.
\newblock \emph{Proteins: Structure}, 48, 2002.

\bibitem[Levy et~al.(2022{\natexlab{a}})Levy, Poitevin, Martel, Nashed, Peck, Miolane, Ratner, Dunne, and Wetzstein]{cryoai}
Axel Levy, Fr{\'e}d{\'e}ric Poitevin, Julien Martel, Youssef Nashed, Ariana Peck, Nina Miolane, Daniel Ratner, Mike Dunne, and Gordon Wetzstein.
\newblock Cryoai: Amortized inference of poses for ab initio reconstruction of 3d molecular volumes from real cryo-em images.
\newblock \emph{arXiv preprint arXiv:2203.08138}, 2022{\natexlab{a}}.

\bibitem[Levy et~al.(2022{\natexlab{b}})Levy, Wetzstein, Martel, Poitevin, and Zhong]{cryofire}
Axel Levy, Gordon Wetzstein, Julien Martel, Frederic Poitevin, and Ellen~D Zhong.
\newblock Amortized inference for heterogeneous reconstruction in cryo-em.
\newblock \emph{arXiv preprint arXiv:2210.07387}, 2022{\natexlab{b}}.

\bibitem[Levy et~al.(2024)Levy, Grzadkowski, Poitevin, Vallese, Clarke, Wetzstein, and Zhong]{drgnai}
Axel Levy, Michal Grzadkowski, Frederic Poitevin, Francesca Vallese, Oliver~B Clarke, Gordon Wetzstein, and Ellen~D Zhong.
\newblock Revealing biomolecular structure and motion with neural ab initio {cryo-EM} reconstruction.
\newblock \emph{bioRxiv}, page 2024.05.30.596729, 2024.

\bibitem[Li et~al.(2023)Li, Zhou, Yuan, Ye, and Gu]{cryostar}
Yilai Li, Yi Zhou, Jing Yuan, Fei Ye, and Quanquan Gu.
\newblock Cryostar: Leveraging structural prior and constraints for cryo-em heterogeneous reconstruction.
\newblock \emph{bioRxiv}, 2023.

\bibitem[Liu et~al.(2023)Liu, Zeng, Qin, Li, Zhang, Xu, and Yu]{cryoformer}
Xinhang Liu, Yan Zeng, Yifan Qin, Hao Li, Jiakai Zhang, Lan Xu, and Jingyi Yu.
\newblock Cryoformer: Continuous heterogeneous cryo-em reconstruction using transformer-based neural representations.
\newblock \emph{arXiv preprint arXiv:2303.16254}, 2023.

\bibitem[Lowe(2004)]{lowe2004distinctive}
David~G Lowe.
\newblock Distinctive image features from scale-invariant keypoints.
\newblock \emph{International journal of computer vision}, 60:\penalty0 91--110, 2004.

\bibitem[Luo et~al.(2024)Luo, Zhang, Xu, Wang, and Ma]{OPUS-DSD2}
Zhenwei Luo, Yiqiu Zhang, Gang Xu, Qinghua Wang, and Jianpeng Ma.
\newblock Opus-dsd2: Disentangling dynamics and compositional heterogeneity for cryo-em single particle analysis.
\newblock \emph{bioRxiv}, 2024.

\bibitem[Nakane et~al.(2018)Nakane, Kimanius, Lindahl, and Scheres]{spliceosome2}
Takanori Nakane, Dari Kimanius, Erik Lindahl, and Sjors~HW Scheres.
\newblock Characterisation of molecular motions in cryo-em single-particle data by multi-body refinement in relion.
\newblock \emph{eLife}, 7:\penalty0 e36861, 2018.

\bibitem[Nashed et~al.(2021)Nashed, Poitevin, Gupta, Woollard, Kagan, Yoon, and Ratner]{cryopose}
Youssef~SG Nashed, Fr{\'e}d{\'e}ric Poitevin, Harshit Gupta, Geoffrey Woollard, Michael Kagan, Chun~Hong Yoon, and Daniel Ratner.
\newblock Cryoposenet: End-to-end simultaneous learning of single-particle orientation and 3d map reconstruction from cryo-electron microscopy data.
\newblock In \emph{Proceedings of the IEEE/CVF International Conference on Computer Vision}, pages 4066--4076, 2021.

\bibitem[Plaschka et~al.(2017)Plaschka, Lin, and Nagai]{spliceosome1}
Clemens Plaschka, Pei-Chun Lin, and Kiyoshi Nagai.
\newblock Structure of a pre-catalytic spliceosome.
\newblock \emph{Nature}, 546, 2017.

\bibitem[Pragier and Shkolnisky(2019)]{pragier2019common}
Gabi Pragier and Yoel Shkolnisky.
\newblock A common lines approach for ab initio modeling of cyclically symmetric molecules.
\newblock \emph{Inverse Problems}, 35\penalty0 (12):\penalty0 124005, 2019.

\bibitem[Punjani et~al.(2017)Punjani, Rubinstein, Fleet, and Brubaker]{punjani2017cryosparc}
Ali Punjani, John~L. Rubinstein, David~J. Fleet, and Marcus~A. Brubaker.
\newblock cryosparc: algorithms for rapid unsupervised cryo-em structure determination.
\newblock \emph{Nature Methods}, 14\penalty0 (3):\penalty0 290--296, 2017.

\bibitem[Rohou and Grigorieff(2015)]{ctffind4}
Alexis Rohou and Nikolaus Grigorieff.
\newblock Ctffind4: Fast and accurate defocus estimation from electron micrographs.
\newblock \emph{Journal of structural biology}, 192\penalty0 (2):\penalty0 216--221, 2015.

\bibitem[Ru et~al.(2015)Ru, Chambers, Fu, Tong, Liao, and Wu]{RAG1-RAG2}
Heng Ru, Melissa Chambers, Tian-Min Fu, Alexander Tong, Maofu Liao, and Hao Wu.
\newblock Molecular mechanism of v(d)j recombination from synaptic rag1-rag2 complex structures.
\newblock \emph{Cell}, 163, 2015.

\bibitem[Sarlin et~al.(2019)Sarlin, DeTone, Malisiewicz, and Rabinovich]{sarlin2019superglue}
Paul{-}Edouard Sarlin, Daniel DeTone, Tomasz Malisiewicz, and Andrew Rabinovich.
\newblock Superglue: Learning feature matching with graph neural networks.
\newblock \emph{CoRR}, abs/1911.11763, 2019.

\bibitem[Scheres(2012)]{scheres2012relion}
Sjors~H.W. Scheres.
\newblock Relion: Implementation of a bayesian approach to cryo-em structure determination.
\newblock \emph{Journal of Structural Biology}, 180\penalty0 (3):\penalty0 519--530, 2012.

\bibitem[Scheres et~al.(2007)Scheres, Gao, Valle, Herman, Eggermont, Frank, and Carazo]{scheres2007disentangling}
Sjors~HW Scheres, Haixiao Gao, Mikel Valle, Gabor~T Herman, Paul~PB Eggermont, Joachim Frank, and Jose-Maria Carazo.
\newblock Disentangling conformational states of macromolecules in 3d-em through likelihood optimization.
\newblock \emph{Nature methods}, 4\penalty0 (1):\penalty0 27--29, 2007.

\bibitem[Schonberger and Frahm(2016)]{schonberger2016structure}
Johannes~L Schonberger and Jan-Michael Frahm.
\newblock Structure-from-motion revisited.
\newblock In \emph{Proceedings of the IEEE conference on computer vision and pattern recognition}, pages 4104--4113, 2016.

\bibitem[Shekarforoush et~al.(2022)Shekarforoush, Lindell, Fleet, and Brubaker]{ResMFN}
Shayan Shekarforoush, David~B Lindell, David~J Fleet, and Marcus~A Brubaker.
\newblock Residual multiplicative filter networks for multiscale reconstruction.
\newblock \emph{arXiv preprint arXiv:2206.00746}, 2022.

\bibitem[Shekarforoush et~al.(2025)Shekarforoush, Lindell, Brubaker, and Fleet]{cryospin}
Shayan Shekarforoush, David Lindell, Marcus~A Brubaker, and David~J Fleet.
\newblock Cryospin: Improving ab-initio cryo-em reconstruction with semi-amortized pose inference.
\newblock \emph{Advances in Neural Information Processing Systems}, 37:\penalty0 55785--55809, 2025.

\bibitem[Shen et~al.(2024)Shen, Dai, Chen, Zeng, Zhang, Pei, and Yu]{draco}
Yingjun Shen, Haizhao Dai, Qihe Chen, Yan Zeng, Jiakai Zhang, Yuan Pei, and Jingyi Yu.
\newblock Draco: A denoising-reconstruction autoencoder for cryo-em.
\newblock \emph{Advances in Neural Information Processing Systems}, 38, 2024.

\bibitem[Singer et~al.(2010)Singer, Coifman, Sigworth, Chester, and Shkolnisky]{singer2010detecting}
Amit Singer, Ronald~R Coifman, Fred~J Sigworth, David~W Chester, and Yoel Shkolnisky.
\newblock Detecting consistent common lines in cryo-em by voting.
\newblock \emph{Journal of structural biology}, 169\penalty0 (3):\penalty0 312--322, 2010.

\bibitem[Tang et~al.(2007)Tang, Peng, Baldwin, Mann, Jiang, Rees, and Ludtke]{EMAN2}
Guang Tang, Liwei Peng, Philip~R. Baldwin, Deepinder~S. Mann, Wen Jiang, Ian Rees, and Steven~J. Ludtke.
\newblock Eman2: An extensible image processing suite for electron microscopy.
\newblock \emph{Journal of Structural Biology}, 157\penalty0 (1):\penalty0 38--46, 2007.
\newblock Software tools for macromolecular microscopy.

\bibitem[Vulović et~al.(2013)Vulović, Ravelli, {van Vliet}, Koster, Lazić, Lücken, Rullgård, Öktem, and Rieger]{cryo-em-formation}
Miloš Vulović, Raimond~B.G. Ravelli, Lucas~J. {van Vliet}, Abraham~J. Koster, Ivan Lazić, Uwe Lücken, Hans Rullgård, Ozan Öktem, and Bernd Rieger.
\newblock Image formation modeling in cryo-electron microscopy.
\newblock \emph{Journal of Structural Biology}, 183\penalty0 (1):\penalty0 19--32, 2013.

\bibitem[Walls et~al.(2020)Walls, Park, Tortorici, Wall, McGuire, and Veesler]{spike}
Alexandra~C. Walls, Young-Jun Park, M.~Alejandra Tortorici, Abigail Wall, Andrew~T. McGuire, and David Veesler.
\newblock Structure, function, and antigenicity of the sars-cov-2 spike glycoprotein.
\newblock \emph{Cell}, 181\penalty0 (2):\penalty0 281--292.e6, 2020.

\bibitem[Wang and Agapito(2024)]{spann3r}
Hengyi Wang and Lourdes Agapito.
\newblock 3d reconstruction with spatial memory.
\newblock \emph{arXiv preprint arXiv:2408.16061}, 2024.

\bibitem[Wang et~al.(2021)Wang, Wang, Peralta, Yaseen, and Pavletich]{fa}
Shengliu Wang, Renjing Wang, Christopher Peralta, Ayat Yaseen, and Nikola~P. Pavletich.
\newblock Structure of the fa core ubiquitin ligase closing the id clamp on dna.
\newblock \emph{Nature Structural \& Molecular Biology}, 28:\penalty0 300 -- 309, 2021.

\bibitem[Wang et~al.(2024)Wang, Leroy, Cabon, Chidlovskii, and Revaud]{wang2024dust3r}
Shuzhe Wang, Vincent Leroy, Yohann Cabon, Boris Chidlovskii, and Jerome Revaud.
\newblock Dust3r: Geometric 3d vision made easy.
\newblock In \emph{Proceedings of the IEEE/CVF Conference on Computer Vision and Pattern Recognition}, pages 20697--20709, 2024.

\bibitem[Weinzaepfel et~al.(2023)Weinzaepfel, Lucas, Leroy, Cabon, Arora, Br{\'e}gier, Csurka, Antsfeld, Chidlovskii, and Revaud]{croco_v2}
Philippe Weinzaepfel, Thomas Lucas, Vincent Leroy, Yohann Cabon, Vaibhav Arora, Romain Br{\'e}gier, Gabriela Csurka, Leonid Antsfeld, Boris Chidlovskii, and J{\'e}r{\^o}me Revaud.
\newblock {CroCo v2: Improved Cross-view Completion Pre-training for Stereo Matching and Optical Flow}.
\newblock In \emph{ICCV}, 2023.

\bibitem[{Weinzaepfel, Philippe and Leroy, Vincent and Lucas, Thomas and Br\'egier, Romain and Cabon, Yohann and Arora, Vaibhav and Antsfeld, Leonid and Chidlovskii, Boris and Csurka, Gabriela and Revaud J\'er\^ome}(2022)]{croco}
{Weinzaepfel, Philippe and Leroy, Vincent and Lucas, Thomas and Br\'egier, Romain and Cabon, Yohann and Arora, Vaibhav and Antsfeld, Leonid and Chidlovskii, Boris and Csurka, Gabriela and Revaud J\'er\^ome}.
\newblock {CroCo: Self-Supervised Pre-training for 3D Vision Tasks by Cross-View Completion}.
\newblock In \emph{{NeurIPS}}, 2022.

\bibitem[Yang et~al.(2025)Yang, Sax, Liang, Henaff, Tang, Cao, Chai, Meier, and Feiszli]{fast3r}
Jianing Yang, Alexander Sax, Kevin~J. Liang, Mikael Henaff, Hao Tang, Ang Cao, Joyce Chai, Franziska Meier, and Matt Feiszli.
\newblock Fast3r: Towards 3d reconstruction of 1000+ images in one forward pass.
\newblock In \emph{Proceedings of the IEEE/CVF Conference on Computer Vision and Pattern Recognition (CVPR)}, 2025.

\bibitem[Zhang et~al.(2024{\natexlab{a}})Zhang, Chen, Zeng, Gao, He, Liu, and Yu]{cryogem}
Jiakai Zhang, Qihe Chen, Yan Zeng, Wenyuan Gao, Xuming He, Zhijie Liu, and Jingyi Yu.
\newblock Cryogem: Physics-informed generative cryo-electron microscopy.
\newblock In \emph{Proceedings of the 38th International Conference on Neural Information Processing Systems}, 2024{\natexlab{a}}.

\bibitem[Zhang et~al.(2024{\natexlab{b}})Zhang, Herrmann, Hur, Jampani, Darrell, Cole, Sun, and Yang]{zhang2024monst3r}
Junyi Zhang, Charles Herrmann, Junhwa Hur, Varun Jampani, Trevor Darrell, Forrester Cole, Deqing Sun, and Ming-Hsuan Yang.
\newblock Monst3r: A simple approach for estimating geometry in the presence of motion.
\newblock \emph{arXiv preprint arxiv:2410.03825}, 2024{\natexlab{b}}.

\bibitem[Zhong et~al.(2021{\natexlab{a}})Zhong, Bepler, Berger, and Davis]{cryodrgn}
Ellen~D Zhong, Tristan Bepler, Bonnie Berger, and Joseph~H Davis.
\newblock Cryodrgn: reconstruction of heterogeneous cryo-em structures using neural networks.
\newblock \emph{Nature methods}, 18\penalty0 (2):\penalty0 176--185, 2021{\natexlab{a}}.

\bibitem[Zhong et~al.(2021{\natexlab{b}})Zhong, Lerer, Davis, and Berger]{cryodrgn2}
Ellen~D. Zhong, Adam Lerer, Joseph~H. Davis, and Bonnie Berger.
\newblock Cryodrgn2: Ab initio neural reconstruction of 3d protein structures from real cryo-em images.
\newblock In \emph{Proceedings of the IEEE/CVF International Conference on Computer Vision (ICCV)}, pages 4066--4075, 2021{\natexlab{b}}.

\end{thebibliography}
}

\end{document}